\documentclass[10pt,twocolumn,letterpaper]{article}

\usepackage{3dv}
\usepackage{times}
\usepackage{epsfig}
\usepackage{graphicx}
\usepackage{amsmath}
\usepackage{amssymb}

\usepackage{enumitem}
\usepackage{multirow}
\usepackage{makecell}
\usepackage{float}
\usepackage[dvipsnames]{xcolor}

\usepackage[pagebackref=true,breaklinks=true,letterpaper=true,colorlinks,bookmarks=false]{hyperref}

\threedvfinalcopy 

\def\threedvPaperID{0220} 

\newcommand{\x}{\mathbf{x}}
\newcommand{\y}{\mathbf{y}}

\DeclareMathAlphabet\mathbfcal{OMS}{cmsy}{b}{n}
\DeclareMathAlphabet\mathbfcal{OMS}{cmsy}{b}{n}

\ifthreedvfinal\fi
\begin{document}

\title{Depth-Aware Action Recognition: Pose-Motion Encoding through Temporal Heatmaps}
\author{Mattia Segu, Federico Pirovano, Gianmario Fumagalli, Amedeo Fabris\\
ETH Zurich\\
{\tt\small \{segum, fpirovan, gfumagal, afabris\} @ethz.ch}
}

\maketitle

\begin{abstract}
    Most state-of-the-art methods for action recognition rely only on 2D spatial features encoding appearance, motion or pose.
    However, 2D data lacks the depth information, which is crucial for recognizing fine-grained actions.
    In this paper, we propose a depth-aware volumetric descriptor that encodes pose and motion information in a unified representation for action classification in-the-wild.
    Our framework is robust to many challenges inherent to action recognition, e.g. variation in viewpoint, scene, clothing and body shape.
    The key component of our method is the Depth-Aware Pose Motion representation (DA-PoTion), a new video descriptor that encodes the 3D movement of semantic keypoints of the human body.
    Given a video, we produce human joint heatmaps for each frame using a state-of-the-art 3D human pose regressor and we give each of them a unique color code according to the relative time in the clip.
    Then, we aggregate such 3D time-encoded heatmaps for all human joints to obtain a fixed-size descriptor (DA-PoTion), which is suitable for classifying actions using a shallow 3D convolutional neural network (CNN).
    The \emph{DA-PoTion} alone defines a new state-of-the-art on the Penn Action Dataset.
    Moreover, we leverage the intrinsic complementarity of our pose motion descriptor with appearance based approaches by combining it with Inflated 3D ConvNet (I3D) to define a new state-of-the-art on the JHMDB Dataset.
\end{abstract}

\section{Introduction} \label{introduction}
Recognizing actions performed by humans is an evergreen problem for the computer vision community.
Action recognition involves identification of different activities from video clips.
The classification of human actions presents applications in several fields, \eg entertainment, video retrieval, and also many safety-critical tasks such as human-robot interaction and video surveillance~\cite{ciptadi2014movement, Akkaladevi2015ActionRF, ji20123d}.

Currently, most state-of-the-art methods for action recognition are 2D-based, and can rely on appearance, optical flow or pose features~\cite{sevilla2018integration, carreira2017quo, liu2019joint, simonyan2014two, Luvizon_2018_CVPR}.
Carreira \etal~\cite{carreira2017quo} recently proposed a two-stream CNN approach~\cite{simonyan2014two} that jointly considers both appearance and motion, obtaining state-of-the-art performance.
Their method processes appearance information using RGB data, while motion is taken into account by leveraging optical flow images. 

Furthermore, pose features have also shown to be effective in recognizing actions~\cite{choutas:hal-01764222, Cheron_2015_ICCV, Luvizon_2018_CVPR}. 
Such techniques first estimate body poses from single images or frame sequences, which are then fed as input to action recognition models. 
\begin{figure}[t]\label{fig:Framework}
\centering
\includegraphics[width = 0.5\textwidth]{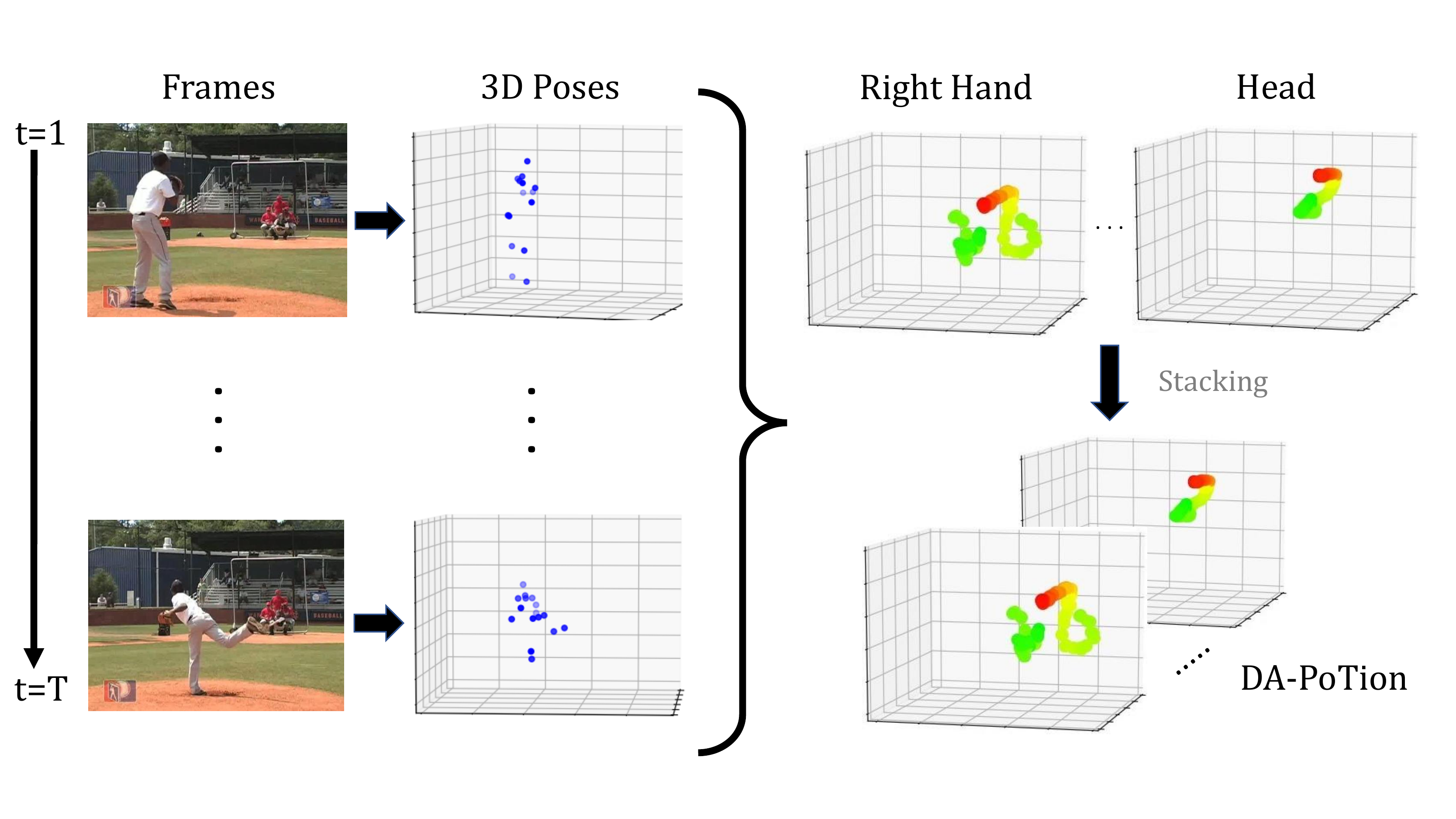}
\centering
\caption{An example of our depth-aware PoTion representation. Given a video, it encodes the 3D movement of semantic keypoints of the human body in a volumetric descriptor.}
\label{img:Framework}
\end{figure}
However, these methods mostly rely on 2D pose information and cannot always achieve the same robustness and accuracy as human vision.
Indeed, many action recognition datasets feature both the same action appearing in several different viewpoints and different actions having the same 2D projection.
Consequently, 2D pose action recognition methods are bound to fail the classification task in these situations.
The Depth-Aware Pose Motion representation we propose, on the other hand, being endowed with depth-awareness, can overcome this crucial limitation and achieve high classification accuracy also for fine-grained actions.

Moreover, most of the popular techniques for action classification generally require pre-trained neural networks~\cite{carreira2017quo, gao2018im2flow, Cheron_2015_ICCV, simonyan2014two}.
In our work, we manage to overcome this limitation, since our proposed video descriptor is \emph{compact} and \emph{fixed-size}, and we directly use it as input to train a CNN \emph{from scratch} for action recognition.

By relying only on 3D pose features, the framework we propose is particularly suitable for solving problems that arise from cluttered scenes and variations in viewpoint, clothing and body shape.
However, as can be inferred from Figure~\ref{img:confusion}, a potential drawback of just considering pose features is that actions involving similar 3D movements, \eg \emph{tennis serve} and \emph{baseball pitch}, can be misclassified.
Nevertheless, we overcome this issue thanks to the complementarity of our framework with appearance-based models, such as the work by Carreira \etal~\cite{carreira2017quo}.
In fact, by combining our DA-PoTion with their Inflated 3D Convnet architecture, we define a new state-of-the-art on the JHMDB Dataset~\cite{Jhuang_2013_ICCV}.

We evaluate our method on Penn-Action~\cite{zhang2013actemes} and \mbox{JHMDB} datasets~\cite{Jhuang_2013_ICCV}, outperforming the baseline provided by the 2D pose motion descriptor~\cite{choutas:hal-01764222} and competing with other state-of-the-art methods.
\section{Related Work}
\subsection{Pose Estimation}
While 2D pose estimation algorithms can now achieve human-like precision~\cite{martinez2018investigating}, 3D pose recovery is still an open research topic, and has gained major interest in the past few years.
Furthermore, as it is difficult to build an in-the-wild dataset with ground-truth 3D annotations~\cite{von2018recovering}, most of the datasets are recorded in lab-like controlled environments.
As a consequence, many 3D pose estimators suffer from low transferability to unseen domains.

Pavlakos \etal~\cite{pavlakos2017coarse}, on the other hand, approach the task with a unified architecture, derived from the `stacked hourglass network'~\cite{newell2016stacked}, that produces a volumetric output featuring per-voxel likelihood for each joint.
Zhou \etal~\cite{zhou2017towards} developed the architecture that we use in this paper.
They specifically tackle the topic of 3D pose retrieval in-the-wild, proposing a weakly-supervised transfer learning approach that uses mixed 2D and 3D labels in a unified deep-learning architecture.
Their main contribution is a 3D pose regressor which can be trained end-to-end with both 2D-annotated in-the-wild images and in-the-lab images with 3D annotations.
\subsection{Action Recognition}
We survey four possible approaches towards action recognition.
At first we examine action recognition methods not involving deep learning, and then we consider deep learning approaches leveraging respectively video features, pose features, and both video and pose features combined.\\\\
{\noindent\textbf{Before deep learning}\quad}
Early action recognition frameworks relied mainly on hand-crafted features~\cite{Kong2018}, such as bag-of-words~\cite{schuldt2004recognizing, dollar2005behavior} or action shape~\cite{blank2005actions}.
Such specific features allow the direct learning of action assignments through simple classifiers, such as Support Vector Machines~\cite{schuldt2004recognizing, laptev2008learning, marszalek2009actions} or K-Nearest Neighbors~\cite{blank2005actions, tran2008human}.
Wang ~\etal~\cite{wang2013dense} introduced a bag-of-features descriptor consisting of `Histograms of Oriented Gradients', `Optical Flow' and `Motion Boundaries' to represent dense trajectories, which are then classified into actions using a kernelized-SVM.
Other methods handle videos as a composition of temporal segments, and train sequential state models, like Hidden Markov Models~\cite{duong2005activity}, Conditional Random Fieds~\cite{wang2006hidden} and Structured SVM ~\cite{niebles2010modeling, tang2012learning} to learn the evolution of human pose and appearance throughout the video.
Despite achieving promising results, all these methods rely on ad-hoc features that require intensive pre-processing and domain specific knowledge~\cite{Kong2018}.
Hence, latest research works have shifted their focus towards deep learning techniques, since neural networks can extract task-dependent features autonomously and in an efficient way~\cite{zeiler2014visualizing}. 
\\\\
{\noindent\textbf{Video features}\quad}
Deep learning models that rely on video features process only appearance and motion information.
Appearance data is given by the raw RGB frames, whereas motion information is obtained by computing optical flow on the video.

The `two-stream architecture' is a very influential model proposed by Simonyan \etal~\cite{simonyan2014two}.
The model they propose includes one spatial stream network that processes individual RGB frames and one temporal stream network which processes multiple optical flow maps.
In the approach proposed by Gao \etal~\cite{gao2018im2flow}, a deep learning model learns to regress the optical flow map of future movements from a single RGB image.
Then, the RGB frame and the learned motion map are combined as input to a classifier for action recognition.
Fergus \etal~\cite{C3D2}, on the other hand, propose a simple approach towards spatio-temporal convolutions over a video.
After learning spatio-temporal features for each video using a 3D-CNN, they use a naive linear classifier for action recognition.
Finally, Carreira \etal~\cite{carreira2017quo} propose the Inflated 3D ConvNet architecture.
In I3D, filters and pooling kernels of deep image classification networks are given a third dimension to learn spatio-temporal features for each video.
The authors propose to train two separate I3D models, one for RGB appearance information and one for smooth optical flow maps.
The resulting scores of the two networks are averaged at test time.\\\\
{\noindent \textbf{Pose features}\quad}
Another category of approaches towards action recognition consists in only considering pose features.

Iqbal \etal~\cite{iqbal2017pose} propose a recursive approach for tackling both pose estimation and action recognition.
Their method is twofold and iterates over a loop, since it learns to predict pose estimates from the distribution of actions and vice versa at the same time.
Finally, Choutas \etal~\cite{choutas:hal-01764222} propose the pose motion (PoTion) descriptor, a fixed-size 2D representation that jointly encodes pose and motion of human joints.
After extracting 2D pose heatmaps for all human joints in a video, they temporally aggregate them to obtain a compact descriptor that can be classified into actions using a shallow CNN.\\\\
{\noindent \textbf{Pose and video features}\quad}
The combination of pose and video features is often the method of choice for action recognition, because it allows to take the best out of the two approaches.

Liu \etal~\cite{liu2019joint} propose two descriptors, the appearance-based Dynamic Texture Image (DTI) and the pose-based Dynamic Pose Image (DPI).
The DTI is a pose-guided video frame feature obtained via an attention-based mechanism.
The DPI representation is the sequence of joint estimation heatmaps.
In contrast to our DA-PoTion, the DPI has many drawbacks: first, it is not fixed-size, as the length of the sequence depends on the length of the clip, second, it does not encode smooth time-encoded trajectories, and third, it does not leverage 3D pose information.
Luvizon \etal~\cite{Luvizon_2018_CVPR} propose a unified framework for 2D/3D pose estimation and action recognition.
Features extracted from the RGB frames are firstly fused with joint heatmaps to provide appeareance recognition, which is then combined with the 2D/3D estimated poses to yield the action prediction.
As will be further discussed in Section \ref{results}, even though their proposed framework includes 3D pose predictions, it does not encode them into a volumetric representation as we do in the DA-PoTion.
Rather, it encodes information in a sparse, scarcely-correlated manner.

The Pose-CNN descriptor proposed by ~\cite{Cheron_2015_ICCV}, on the other hand, aggregates motion and appearance information along directions of human joints.
The proposed model leverages pose information to guide the attention of RGB and optical flow feature extractors.
\section{Method}
In this Section we present our end-to-end 3D action recognition framework.
Our method is divided into three stages:
given a video,
\begin{enumerate}[label=(\roman*)]
    \item 
    we extract human joint heatmaps for each frame;
    \item
    we assign a unique color code to each joint heatmap depending on the relative time in the video clip, and we aggregate them to obtain a fixed-size descriptor, the DA-PoTion. The resulting representation gracefully fuses pose and motion information and summarizes the whole video;
    \item
    we use these descriptors to train from scratch a shallow \mbox{3D-CNN} for action classification.
\end{enumerate}
\subsection{3D Pose Extraction}\label{section_1}
In the first stage of our framework, we regress 3D joint positions for a single subject in the scene.
To accomplish this task, we use the state-of-the-art weakly-supervised algorithm proposed by Zhou \etal~\cite{zhou2017towards}, as we found it to generalize much better to in-the-wild images, if compared to other tested approaches~\cite{pavlakos2017coarse, Zhou2018Starmap}.
We chose not to fine-tune the pose estimator on each dataset, since we wanted our framework to be as general as possible.
The algorithm takes as input a still image and a bounding-box around the human subject and returns predicted 3D positions for \mbox{$J=16$} human joints. 
The 2D $x$ and $y$ coordinates refer to the bounding-box coordinate frame, whereas the depth $z$ coordinate refers to the absolute camera frame. Each coordinate is discretized into $256$ values.

When dealing with a dataset which does not provide bounding box annotations, or with in-the-wild videos, we predict a bounding-box around the subject using YOLOv3~\cite{redmon2018yolov3}. This is often the case for standard action recognition datasets, such as HMDB~\cite{Jhuang_2013_ICCV} or UCF-101~\cite{soomro2012ucf101}.
\subsection{DA-PoTion}\label{section_2}
To represent each video, we propose a compact \mbox{volumetric} descriptor that embodies the 3D trajectories of human joints. Unlike the representation proposed by Choutas \etal~\cite{choutas:hal-01764222}, our descriptor for action recognition is endowed with depth-awareness.
Hence, we call it Depth-Aware Pose Motion (\emph{DA-PoTion}) representation.

In the following, we will describe the process to create the DA-PoTion \emph{for a single human joint}. 
Denoting by $T$ the number of frames in a video, we first convert the $T$ predicted 2D coordinates $(\x,\y)$ from the \mbox{bounding-box} coordinate frame into the coordinate frame of the original image.
Such transformation is achieved by rescaling their values into the size of the original image and then summing the coordinates of the upper-left corner of the bounding-box of each frame, $(\x_{ul}, \y_{ul})$.
By doing this operation, the \mbox{\emph{DA-PoTion}} can encode movements relative to all the camera frame, and its field of view is not restricted to just the motion happening inside the bounding-box.
The immediate consequence of this design choice is that the \emph{DA-PoTion} can distinguish between fine-grained actions such as running and running on the spot.

Then, for each frame, we convert the $T$ predicted point-estimates for the 3D positions into volumetric heatmaps.
This is achieved by applying to a $W\times H\times D$ volume a 3D multivariate Gaussian centered at the point-estimates, where $W$, $H$ and $D$ stand for width, height and depth respectively.
We denote each of these synthetic heatmaps for a given joint $j$ as $\mathcal{H}^t_j[x,y,z]$, for $t \in \{1, 2, \cdots, T\}$.

\begin{figure}[t]\label{fig:Col_scheme}
    \centering
    \includegraphics[width = 0.5\textwidth]{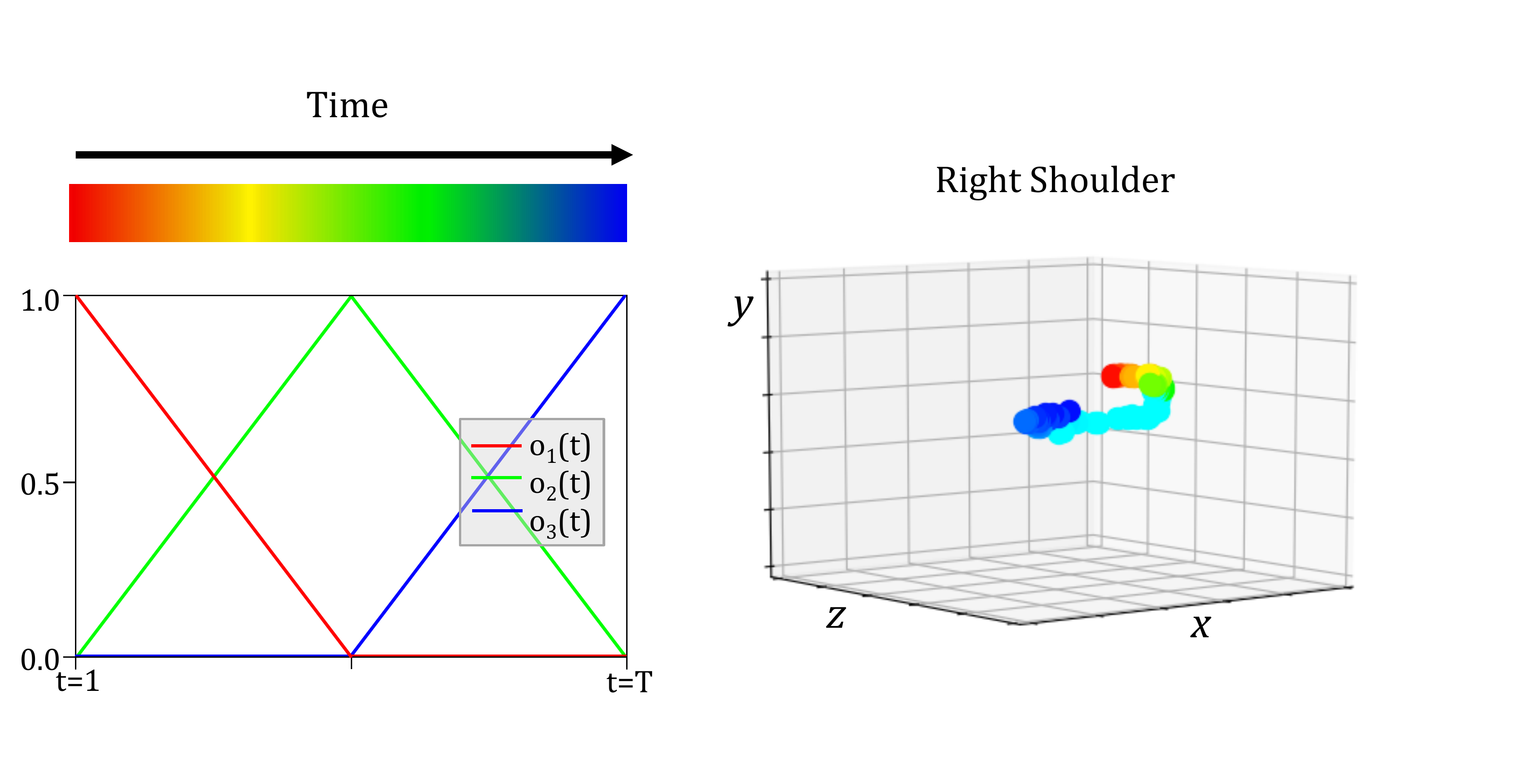}
    \caption{Illustration of the colorization scheme for $C = 3$ (left) and an example of the resulting DA-PoTion representation (right).}
    \label{img:Col_scheme}
\end{figure}

These heatmaps, then, undergo what we call a colorization process, yielding $T$ colorized heatmaps $\mathcal{C}^t_j[x,y,z,c]$ of dimension $W\times H \times D \times C$.
We colorize the heatmaps by assigning to each of the $\mathcal{H}_j^t$ a unique code using $C$ channels, taking inspiration from the 2D-PoTion descriptor~\cite{choutas:hal-01764222}.
For each $t$, we compute a code vector $o(t) \in \mathbb{R}^C$. For values of $C$ up to $3$, we can think $o(t)$ to represent a color vector with $C$ channels. An example of colorization scheme is depicted in Figure~\ref{img:Col_scheme}.
For $C=2$, $o(t)$ is defined as follows:
\begin{equation} \label{eq::color_scheme}
    o(t) = \left(  \frac{t-1}{T-1}, 1 - \frac{t-1}{T-1} \right)^\top.
\end{equation}
The extension to a generic number of channels can be done by splitting the $T$ frames into $C-1$ equally sampled intervals. For each of these intervals, we apply the colorization scheme with $2$ channels out of $C$ total.
The final colorized heatmap for frame $t$ is given by:
\begin{equation}\label{eq_C}
    \mathcal{C}_j^t[x,y,z,c] = \mathcal{H}_j^t[x,y,z] \cdot o_c(t).
\end{equation}
For a given joint $j$, the colorized heatmaps are then summed up over time, yielding a $C$-channel volume $\mathcal{S}_j$, whose size does not depend on the length of the video:
\begin{equation}\label{eq_S}
    \mathcal{S}_j[x,y,z,c] =\sum_{t=1}^{T} \mathcal{C}_j^t[x,y,z,c].
\end{equation}
Still considering a single joint, we now propose three different aggregation schemes, yielding three different versions of the DA-PoTion.

Since the values of $\mathcal{S}_j$ are obtained by summing over time, they depend on the number of frames. Therefore, we normalize each channel $c \in \{1,2, \cdots, C\}$ independently dividing by the maximum value over all voxels.
This operation defines the first, standard aggregation scheme, which we will call \emph{DA Unnormalized PoTion}. It is denoted by $\mathcal{U}_j[x,y,z,c]$ and it is computed as follows:
\begin{equation}\label{eq_U}
    \mathcal{U}_j[x,y,z,c] = \dfrac{\mathcal{S}_j[x,y,z,c]}{\max_{x',y',z'}\mathcal{S}_j[x',y',z',c]}.
\end{equation}
\begin{figure}[t]
\begin{center}
    \includegraphics[scale=0.14]{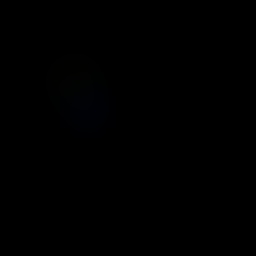} 
    \includegraphics[scale=0.14]{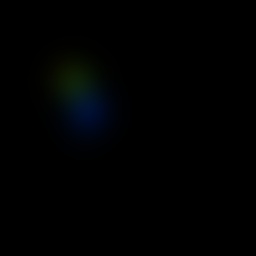} 
    \includegraphics[scale=0.14]{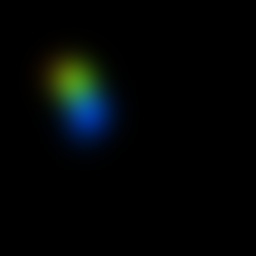} 
    \includegraphics[scale=0.14]{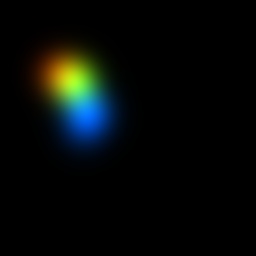} 
    \includegraphics[scale=0.14]{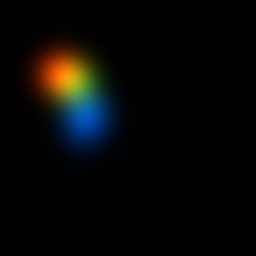} 
    \includegraphics[scale=0.14]{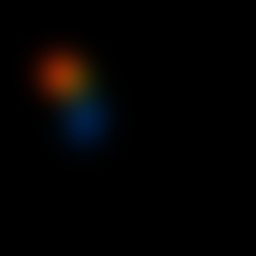} 
    \includegraphics[scale=0.14]{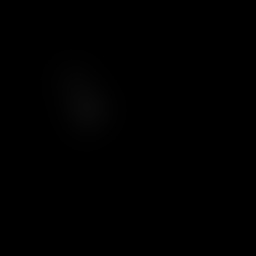} 
    \includegraphics[scale=0.14]{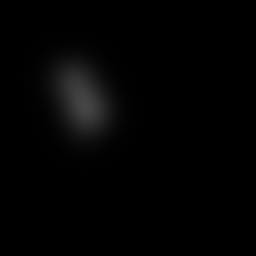} 
    \includegraphics[scale=0.14]{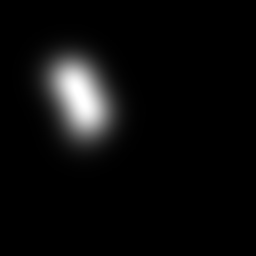} 
    \includegraphics[scale=0.14]{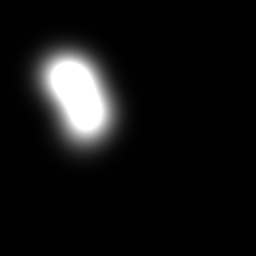} 
    \includegraphics[scale=0.14]{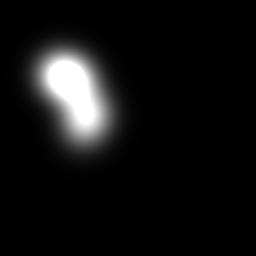} 
    \includegraphics[scale=0.14]{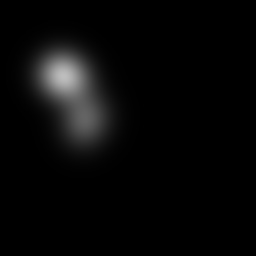} 
    \includegraphics[scale=0.14]{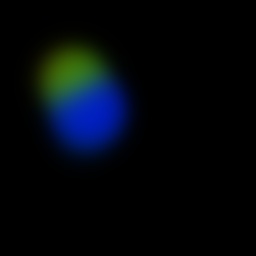} 
    \includegraphics[scale=0.14]{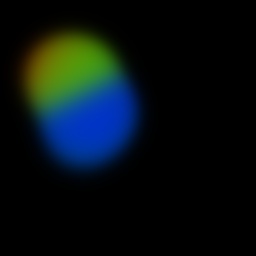} 
    \includegraphics[scale=0.14]{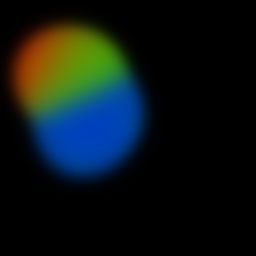} 
    \includegraphics[scale=0.14]{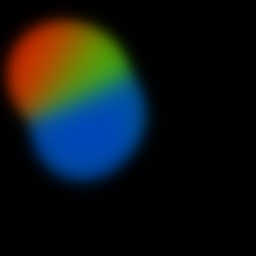} 
    \includegraphics[scale=0.14]{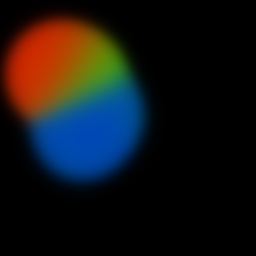} 
    \includegraphics[scale=0.14]{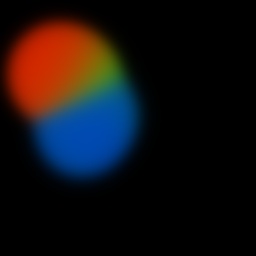} 
\end{center}
    \caption{Examples of DA-PoTion representations for a single human joint. From left to right: increasing levels of depth. Top row: $\mathcal{U}_j$; middle row: $\mathcal{I}_j$; bottom row: $\mathcal{N}_j$.}
    \label{img:3d_potion_photos}
\end{figure}
Figure~\ref{img:3d_potion_photos} shows, in the top row, an example of $\mathcal{U}_j$ at increasing depth levels.
Notice how the time information is encoded in the color of the trajectories.
\begin{figure*}[t]
\includegraphics[width = \textwidth]{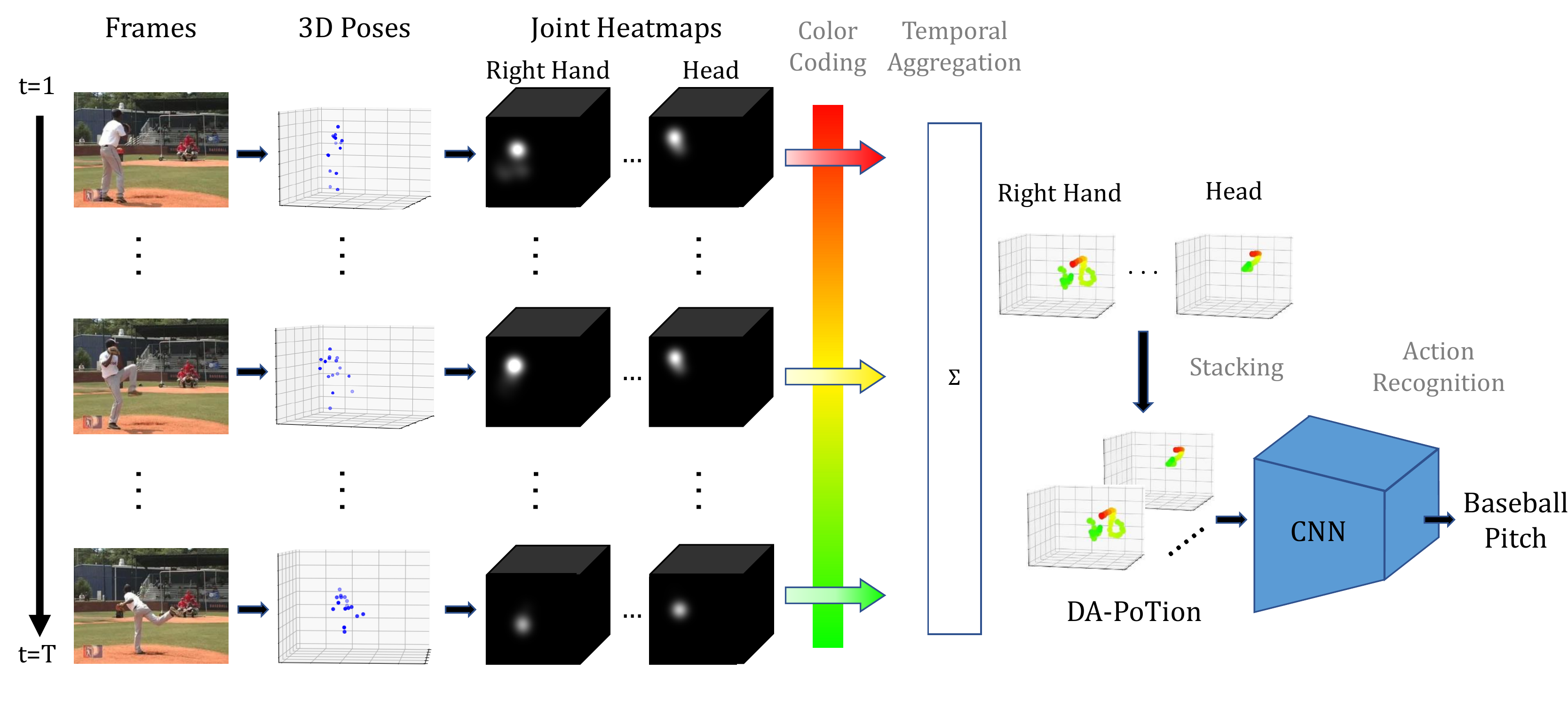}
\caption{Illustration of our end-to-end framework for action recognition (with number of color channels $C=2$). Given a video, (i) we produce 3D human joint heatmaps for each frame; (ii) we assign a unique code to each of them according to the relative time of the frame in the video clip; (iii) we aggregate the 3D time-encoded heatmaps for all human joints to obtain the clip-level, fixed-size DA-PoTion representation. (iv) The resulting descriptor is then fed as input into a shallow 3D-CNN trained for action classification.}
\label{img:Shallow}
\end{figure*}
The second aggregation scheme yields the \emph{DA Intensity PoTion} and it is computed as follows:
\begin{equation}\label{eq_I}
    \mathcal{I}_j[x,y,z] =\sum_{c=1}^{C} \mathcal{U}_j[x,y,z,c].
\end{equation}
Figure~\ref{img:3d_potion_photos} shows, in the middle row, an example of $\mathcal{I}_j$ at increasing depth levels.
If a joint stays at a given 3D position for a non-trivial time, then a stronger intensity will be accumulated in $\mathcal{U}_j$ and $\mathcal{I}_j$. The third aggregation scheme, on the other hand, is defined in such a way to give all positions of the motion trajectory the same weight.

The third aggregation scheme yields the \emph{DA Normalized PoTion} and it is computed as follows:
\begin{equation}\label{eq_N}
    \mathcal{N}_j[x,y,z,c] = \dfrac{\mathcal{U}_j[x,y,z,c]}{\epsilon+\mathcal{I}_j[x,y,z,c]},
\end{equation}
with $\epsilon = 1$ to avoid instabilities in areas with low intensity.
Figure~\ref{img:3d_potion_photos} shows, in the bottom row, an example of $\mathcal{N}_j$ at increasing depth levels.
Notice, in contrast to the top and middle rows of the Figure, how all the steps of the trajectory are weighted equally, regardless of the time spent at each location.

This intrinsic structural difference suggests that the three aggregation schemes convey complementary information and that high classification accuracy can be achieved by stacking $\mathcal{N}$, $\mathcal{U}$ and $\mathcal{I}$ together in what will be referred to as $\mathcal{N}+\mathcal{U}+\mathcal{I}$.

The final DA-PoTion representation for a given aggregation scheme is obtained by stacking the DA-PoTion representations for all human joints.
\subsection{Action Classification} \label{section_3}
Once the DA-PoTion representations for each video in the considered dataset are collected, our framework classifies these descriptors into actions.
Since the DA-PoTions are fixed-size and have significantly less texture than a normal image, we were able to classify them with just a shallow 3D-CNN without any pre-training.\\
Figure~\ref{img:Shallow} illustrates the shallow 3D-CNN architecture that we used to classify the DA-PoTion descriptors.
The input to the network for one clip is the DA-PoTion representation stacked for all human joints.
When using the $\mathcal{U}$ or the $\mathcal{N}$ aggregation scheme, the input is composed of $J\times C$ channels, where $J$ is the number of joints detected by the pose estimator.
When using $\mathcal{I}$, the input is composed of just $J$ channels and, when using the $\mathcal{N}+\mathcal{U}+\mathcal{I}$ aggregation scheme, the input is composed of $J\times \left(  2C+1  \right)$ channels.
Each channel is resized from $W\times H\times D$ to $64\times64\times64$.\\
The network consists of three blocks, each composed of two 3D convolutional layers.
Each convolution has a kernel size of $3\times3\times3$ and is followed by a dropout layer~\cite{srivastava2014dropout} with drop rate $p=0.25$, batch normalization~\cite{ioffe2015batch} and ReLU non-linearity.
The first convolution in each block has a stride of $1$ and the second one has a stride of $2$. Therefore, the spatial resolution of each dimension is halved after each block.
Finally, after the three convolutional blocks, we perform global average pooling followed by a fully-connected layer with soft-max non-linearity to perform video classification.
\\\\
{\noindent \textbf{Training details} \quad}
We initialize all the layers' weights using Xavier initialization~\cite{glorot2010understanding}. Then we optimize the network parameters over 100 epochs using Adam~\cite{kingma2014adam} optimizer, with an exponentially decaying learning rate.
It took approximately 2 days to train our 3D-CNN on a single GPU NVIDIA GeForce GTX 1060.

During training, we use \emph{data augmentation} to improve generalization. Each input sample first goes through a random affine transformation, yielding a translated and rotated output, which is then randomly flipped about the $y$ and $z$ axes.
\subsection{Fusing the DA-PoTion with I3D} \label{section_4}
Since our framework only relies on pose information, its predictions are complementary to the ones produced by I3D~\cite{carreira2017quo} and other approaches that process video features (RGB and optical flow).
For this reason, we implemented in our framework the possibility to fuse the predictions made by the two models.\\
At training time, the two models are trained separately, with the DA-PoTion following the approach described in Sections \ref{section_1},\ref{section_2} and \ref{section_3}, and the I3D model following the procedure reported in ~\cite{carreira2017quo}.
At test time, each video is fed into the DA-PoTion framework, the RGB I3D network and the optical flow I3D network, producing three sets of per-action scores.
We then obtain the final score for each action by averaging the three scores with equal weights.
\section{Results}\label{results}
We ran extensive experiments to validate our approach. Table \ref{N_U_I_3_channels} reports our performance against the 2D PoTion~\cite{choutas:hal-01764222} baseline, Tables \ref{onlyPENN_CHANNELS}, \ref{onlyPENN_TYPE}, and \ref{variance_experiments} present the results of our ablation study and Table \ref{SOTA} presents the comparison of our results with the state-of-the-art.
We tested our end-to-end framework on the \emph{Penn Action}~\cite{zhang2013actemes} and \emph{JHMDB}~\cite{Jhuang_2013_ICCV} datasets. Our results are reported as mean classification accuracy.

The Penn Action Dataset contains $2326$ videos from $15$ classes. It features a single subject per video and it is annotated with action, 2D ground-truth joint positions, and bounding-boxes around the subject.
The JHMDB Dataset contains $928$ videos from $21$ classes. It can feature multiple subjects in the scene. All videos are annotated with 2D ground-truth joint positions for only one subject. The \emph{subJHMDB} Dataset is a subset of JHMDB containing only $12$ actions.\\
%
\begin{table}[t]
    \centering
    \begin{tabular}{ |l||c|c|c|}
    \hline
    Method & Ground-truth & Penn Action\\
    \hline\hline
    2D PoTion~\cite{choutas:hal-01764222} & - & 93.6\\
    2D PoTion~\cite{choutas:hal-01764222} & \checkmark & 95.6\\
    \hline
    \textbf{DA-PoTion} & - & \textbf{96.3}\\
    \hline
    \end{tabular}
\caption{Classification accuracy on the Penn Action Dataset. All experiments were run with $C=3$ channels and $\mathcal{N}+\mathcal{U}+\mathcal{I}$ aggregation scheme. `Ground-truth' denotes that ground-truth pose annotations were used to compute the result.}
\label{N_U_I_3_channels}
\end{table}
\begin{table}[t]
    \centering
    \begin{tabular}{ |l||c|c|}
    \hline
    Method & $\sigma$ & Penn Action\\
    \hline\hline
    DA-PoTion & $2\sqrt{2}$ & 93.6 \\
    DA-PoTion & 4 & 96.3 \\
    DA-PoTion & $4\sqrt{2}$ & 51.7 \\
    \hline
    \end{tabular}
\caption{Classification accuracy on Penn Action Dataset when varying standard deviation $\sigma$ of the Gaussian mask applied on the pose point-predictions. The experiments were run with $C=3$ channels and aggregation scheme $\mathcal{N}+\mathcal{U}+\mathcal{I}.$}
\label{variance_experiments}
\end{table}
\begin{table}[t]
    \centering
    \begin{tabular}{ |l||c|c|}
    \hline
    Method & Channels & Penn Action\\
    \hline\hline
    2D PoTion~\cite{choutas:hal-01764222} & 2 & 90.8 \\
    2D PoTion~\cite{choutas:hal-01764222} & 3 & 93.6 \\
    2D PoTion~\cite{choutas:hal-01764222} & 4 & 94.2 \\
    \hline
    DA-PoTion & 2 & 94.1 \\
    DA-PoTion & 3 & 96.3 \\
    \textbf{DA-PoTion} & \textbf{4} & \textbf{97.2} \\
    \hline
    \end{tabular}
\caption{Classification accuracy on the Penn Action Dataset when varying the number of channels $C$. The experiments were run with $\sigma = 4$ and aggregation scheme $\mathcal{N}+\mathcal{U}+\mathcal{I}$.}
\label{onlyPENN_CHANNELS}
\end{table}

{\noindent \textbf{Baseline comparison}\quad}
As a first experiment, we wanted to compare our performance to the 2D PoTion approach by Choutas \etal~\cite{choutas:hal-01764222}, the baseline to our work.
Table \ref{N_U_I_3_channels} presents our result on the Penn Action Dataset and compares it with the classification accuracy achieved by 2D PoTion.
In order to assess an upper bound for the performance of the 2D PoTion architecture, we ran their framework using the ground-truth annotations of the 2D poses, instead of predicting them.
As we perform better than the 2D PoTion also in this extreme case, we can safely claim that our approach outperforms the baseline.\\\\
{\noindent \textbf{Ablation study} \quad}
In Table \ref{variance_experiments} we explore the effect of changing the variance of the Gaussian mask that is applied to the joint point-predictions extracted by the 3D pose regressor.
In Tables~\ref{onlyPENN_CHANNELS} and ~\ref{onlyPENN_TYPE}, on the other hand, we study the impact of varying number of channels and aggregation schemes in both the 2D PoTion and in our DA-PoTion representation.
\begin{table}[t]
    \centering
    \begin{tabular}{ |l||c|c|}
    \hline
    Method & Type & Penn Action\\
    \hline\hline
    2D PoTion~\cite{choutas:hal-01764222} & $\mathcal{U}$ &  89.7 \\
    2D PoTion~\cite{choutas:hal-01764222} & $\mathcal{N}+\mathcal{U}+\mathcal{I}$ & 93.6\\
    \hline
    DA-PoTion & $\mathcal{U}$ & 95.9 \\
    \textbf{DA-PoTion} & $\mathbfcal{N} + \mathbfcal{U} + \mathbfcal{I}$ & \textbf{96.3} \\
    \hline
    \end{tabular}
\caption{Classification accuracy on the Penn Action Dataset when varying the aggregation scheme. The experiments were run with $\sigma=4$ and $C=3$ channels.}
\label{onlyPENN_TYPE}
\end{table}
\begin{figure*}[t!]
\begin{center}
    \includegraphics[width=0.48\textwidth]{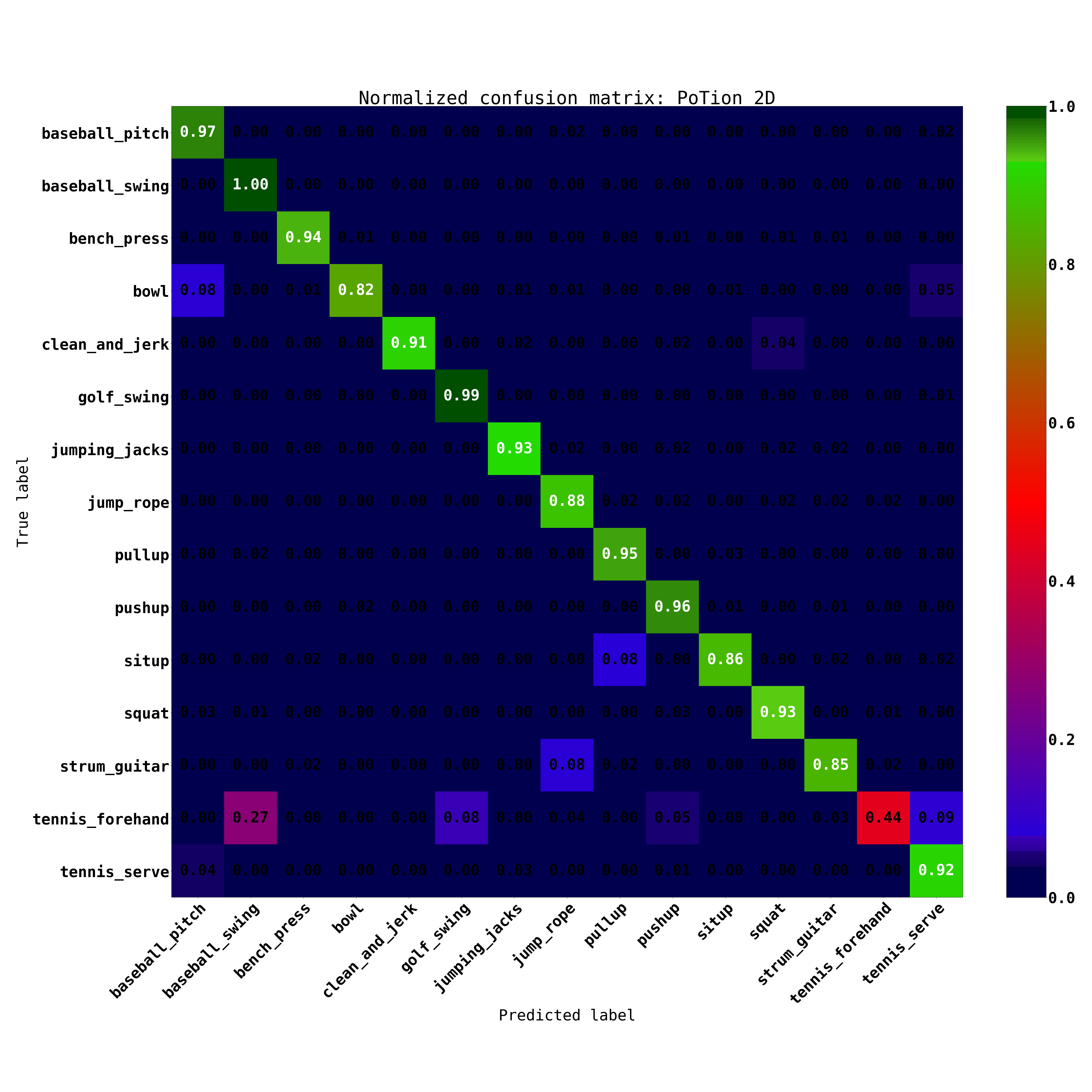} 
    \includegraphics[width=0.48\textwidth]{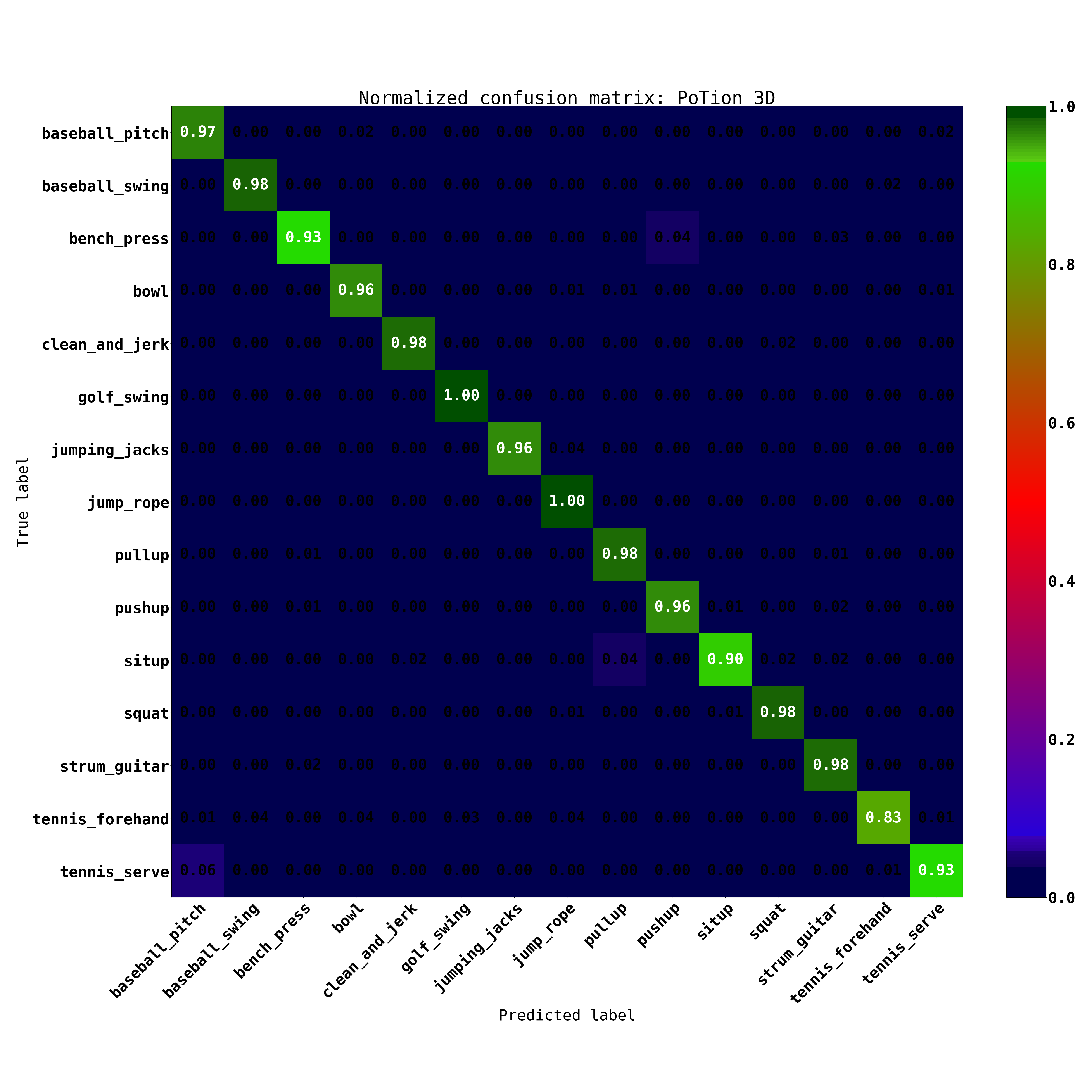} 
\end{center}
    \caption{Confusion matrices for action classification on the Penn Action Dataset. Left: 2D PoTion; right: DA-PoTion.}
    \label{img:confusion}
\end{figure*}

\newcommand{\tabincell}[2]{\begin{tabular}{@{}#1@{}}#2\end{tabular}}
\begin{table}[t!]
\begin{center}
\resizebox{\linewidth}{!}{%
\begin{tabular}{|c|c|c|c|c|}
  \hline
  \multirow{2}{*}{ }
  & \multirow{2}{*}{Method}
  & \multicolumn{3}{c|}{Accuracy}  \\
  \cline{3-5}
  &   &  \tabincell{c}{sub-\\JHMDB}  & \tabincell{c}{JHMDB} &  \tabincell{c}{Penn\\ Action} \\
  \hline
  \hline
  \multirow{5}{*}{\tabincell{c}{Video \\features}}
  & Im2Flow~\cite{gao2018im2flow}                        & - & - & 77.4 \\
  & DT~\cite{wang2013dense}         & 46.0  & -& 73.4 \\
  & C3D~\cite{C3D2}                                  & 68.8  & -& 86.0 \\
   & I3D~\cite{carreira2017quo}      & - & 84.1 & - \\
  \hline
  \hline
  \multirow{5}{*}{\tabincell{c}{Pose \\features}}
   & HLPF \cite{Jhuang_2013_ICCV}                         & 51.1 &- & -\\
   & ACPS \cite{iqbal2017pose}         & 61.5 &- & 79.0  \\
   \cline{2-5}
   & 2D PoTion~\cite{choutas:hal-01764222}                    & - & 57.0 & 93.6 \\
   & \textbf{DA-PoTion} (\textit{Ours})               & \underline{71.5} & \underline{70.4} & \underline{\textbf{97.2}} \\
  \hline
  \hline
  \multirow{9}{*}{\tabincell{c}{Pose+\\Video \\features}}
   & IDT-FV~\cite{wangICCV2013iDT}                         & 60.9  &- & 92.0 \\
   & DTI+DPI~\cite{liu2019joint}
    & - & - & 95.86\\
   & \textbf{Luvizon \etal}~\cite{Luvizon_2018_CVPR}              & - & -& \textbf{97.4*} \\
   & MST-AOG~\cite{DBLPconfcvprWangNXWZ14}                      & 45.3  &- & 74.0 \\
   & ST-AOG~\cite{cvprNieXZ15AndOrGraphsubJHMDBPenn}       & 61.2  & -& 85.5     \\
   & P-CNN~\cite{Cheron_2015_ICCV}                        & 66.8  & 61.1& 95.3 \\
   & Action Tubes~\cite{gkioxari2015finding}                        & -  & 62.5& - \\
   & ACPS+IDT-FV~\cite{iqbal2017pose}  & 74.6  &- & 92.9     \\
   & \textbf{JDD}~\cite{cao2017body}                  & \textbf{79.7}  &- & 95.3 \\
   & Zolfaghari \etal~\cite{Zolfaghari_2017_ICCV}                             & - & 76.1 & -\\
   \cline{2-5}
   & 2D PoTion + I3D~\cite{choutas:hal-01764222}                    & - & 85.5 & - \\
   & \textbf{DA-PoTion} (\textit{Ours}) \textbf{ + I3D}               & - & \textbf{87.8} & - \\
  \hline
\end{tabular}}
\caption{Comparison between our method and the state-of-the-art on the Penn Action, JHMDB and subJHMDB datasets. `Video features' refer to methods using appearance and motion information; `pose features' refer to methods using only pose features. The asterisk denotes that fine-tuning was performed on the pose estimator for the specific dataset. In bold the best method; underlined the best method among those that leverage only pose features.}
\label{SOTA}
\end{center}
\vspace{-6mm}
\end{table}

In Table~\ref{variance_experiments} we highlight the importance of choosing the appropriate variance when creating Gaussian heatmaps from pose point-predictions.
Empirically, we find out that using a variance of $16$ pixels, that is a standard deviation of $4$ pixels, represents the best trade-off between smoothness and sparsity of the representation.
Indeed, when using a smaller variance, the resulting DA-PoTion suffers from noise and discretization errors, and hence it cannot encode smooth, continuous movements of the joints.
This results in a slight degradation of the classification accuracy, as reported in the first row of Table \ref{variance_experiments}.
On the other hand, as the last row of Table \ref{variance_experiments} testifies, when using a bigger variance, the resulting DA-PoTion is almost not informative at all.
In fact, having such a large variance causes the joint trajectories time-encoded in the DA-PoTion to have a thickness that covers almost all the volume of the representation.
Clearly, in such a scenario the 3D-CNN used for action classification cannot distinguish neither coarse-grained nor fine-grained movements.

In Table~\ref{onlyPENN_CHANNELS}, we observe a clear improvement for both the 2D PoTion and DA-PoTion when increasing the number of channels from $2$ to $4$. Nevertheless, a $4$-channel DA-PoTion resulted very computationally expensive. For this reason, for all subsequent results, we report results with $C=3$ as a trade-off between accuracy and compactness of the representation.

In Table~\ref{onlyPENN_TYPE}, we observe that the $\mathcal{N}+\mathcal{U}+\mathcal{I}$ aggregation scheme outperforms the \emph{Unnormalized PoTion} $\mathcal{U}$ for the 2D case by a greater mean than for our method.
Note that the likelihood of having a joint staying at the same 3D position for a non-trivial time, which would lead to differences between $\mathcal{N}$, $\mathcal{U}$ and $\mathcal{I}$, is much lower than the likelihood of the same event happening in the 2D setting.
For the 2D PoTion, stacking the three aggregation schemes conveys additional information that is very useful to distinguish fine-grained actions. This additional information is already intrinsic in our depth-aware representation, even for $\mathcal{U}$.
\\\\
{\noindent \textbf{Comparison to state-of-the-art}\quad}
In Table~\ref{SOTA}, we provide a detailed comparison between our method and other state-of-the-art approaches on the Penn Action, JHMDB, sub-JHMDB datasets. 
We can observe that our framework reaches state-of-the-art accuracy on all three datasets among the methods that only consider pose features. 
Moreover, we define a new state-of-the-art on the Penn Action Dataset~\cite{zhang2013actemes}.
The only result that competes with ours,~\cite{Luvizon_2018_CVPR}, is obtained by fine-tuning its embedded pose regressor on this specific dataset, which we chose not to do in order to improve generalization.

Combining our approach with I3D~\cite{carreira2017quo} in the way described in Section \ref{section_4}, we also define a new state-of-the-art on the JHMDB Dataset~\cite{Jhuang_2013_ICCV}.\\

{\noindent \textbf{Penn Action Dataset performance analysis} \quad}
As can be seen from the confusion matrices in Figure~\ref{img:confusion}, we consistently outperform the 2D PoTion~\cite{choutas:hal-01764222} approach (left image) in all the classes. 
As an example, because of its lack of depth information, 2D PoTion mainly misclassifies \emph{tennis forehand} with \emph{baseball swing} and \emph{golf swing}, which have similar projections on the 2D plane. 
Our approach (right image) is able to make the correct prediction also in this situation by leveraging 3D pose information.
%
\begin{figure}[t]\label{fig:Federer}
    \centering
    \includegraphics[width = 0.45\textwidth]{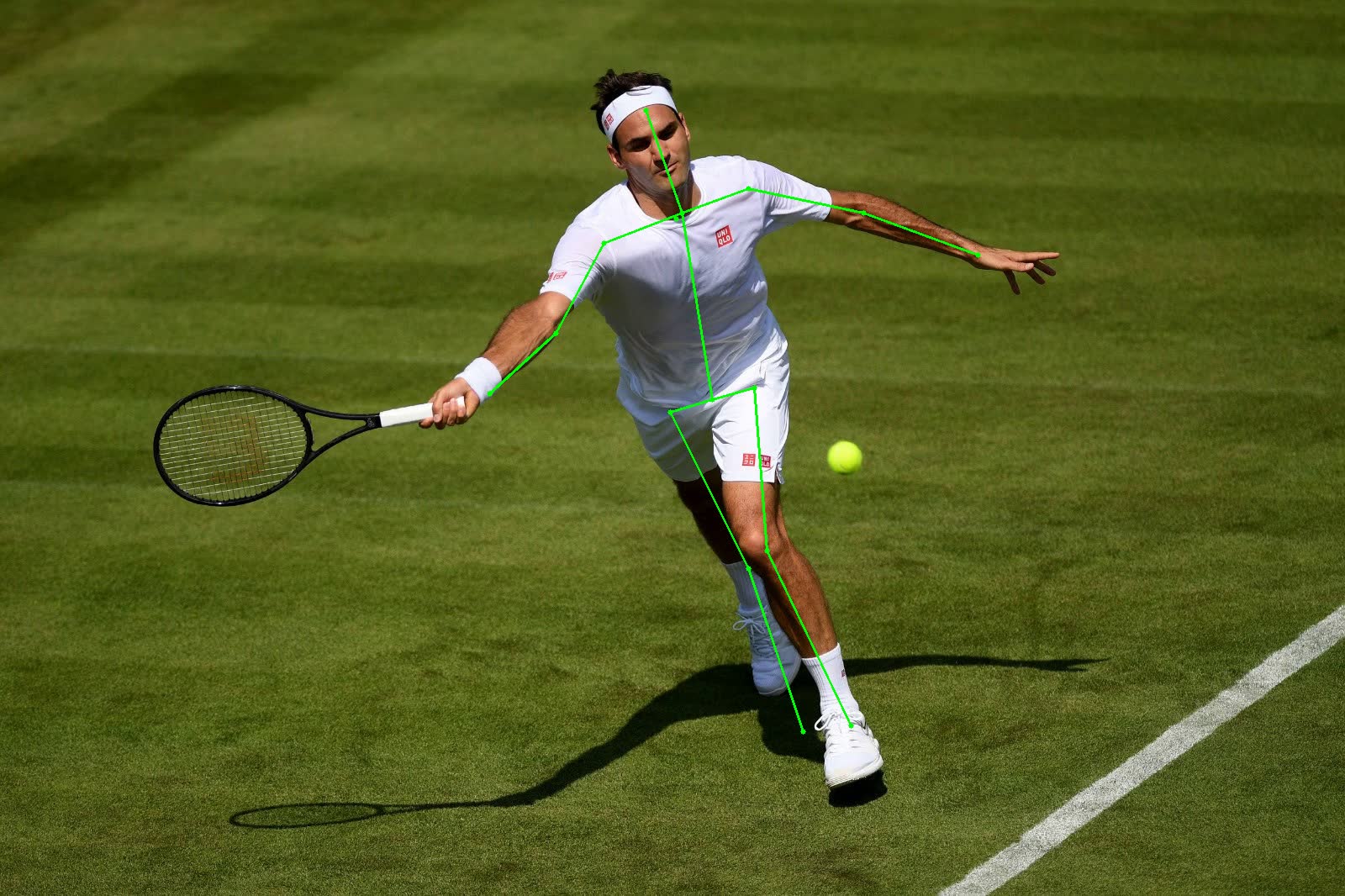}
    \includegraphics[width = 0.45\textwidth]{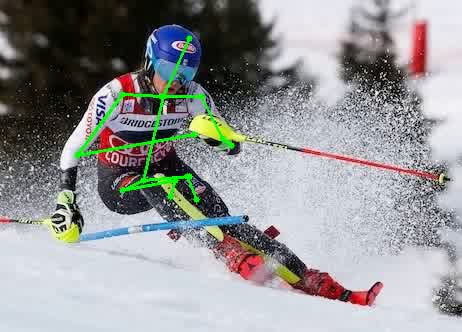}
    \caption{Qualitative performance of the 3D pose estimator by Zhou \etal~\cite{zhou2017towards} on two in-the-wild images. Top: coherent 3D pose predictions; bottom: wrong 3D pose predictions.}
    \label{img:Federer}
\end{figure}\\\\
{\noindent \textbf{Qualitative analysis of the 3D pose estimator} \quad}
It is worth noticing that, despite we are already using a state-of-the-art 3D pose regressor, the potential of our framework can increase even more as 3D pose regressors keep improving. 
In fact, Table~\ref{onlyPENN_TYPE} points out that, when considering ground-truth 2D pose annotations, classification accuracy grows for the 2D setting. This proves that our framework could benefit all the more from improvements of 3D pose regressors.
Indeed, 3D state-of-the-art pose estimators currently perform poorly compared to the 2D ones because of their scarce transferability on domains different than the training one~\cite{martinez2018investigating}.
Low transferability can also be inferred from Figure~\ref{img:Federer}, where we present the qualitative performance of ~\cite{zhou2017towards} on two in-the-wild images.
As the subjects in both images are well included in the frame and well-distinguishable from the background, the main reason for the performance degradation is to be found in the domain shift from the tennis match to the ski race.
\section{Conclusion}
We propose an end-to-end framework for action recognition that reaches state-of-the-art performance, among the methods that rely only on pose features, on the most popular datasets for action classification.
The use of our DA-PoTion representation alone already defines a new state-of-the-art on the Penn Action Dataset~\cite{zhang2013actemes}. Since sport actions often present repetitive patterns for joint movements, which is exactly the key information encoded in our descriptor, we obtain remarkable results on this dataset featuring many sport activities.
Moreover, the combination of our pose feature with the RGB and optical flow streams of I3D~\cite{carreira2017quo} achieves the new state-of-the-art on the JHMDB dataset~\cite{Jhuang_2013_ICCV}, proving how the complementarity of pose and video features can be beneficial to action recognition techniques.
In contrast to the work of Luvizon~\etal~\cite{Luvizon_2018_CVPR}, which also uses 3D poses for action recognition, our DA-PoTion representation is an image-like 3D heatmap whose adjacent points are highly correlated and, thus, suggests the use of a 3D-CNN to learn action assignments from such a dense representation.
Our approach neatly embodies strongly correlated features in a compact descriptor, whereas Luvizon \etal propose the use of a sparse 3D representation with scarcely correlated adjacent points.
Since their descriptor presents high correlation only on the temporal axis, they are forced to use a 2D-CNN.

Additionally, being succint and dense with information, our DA-PoTion is ideal for tasks where the simplicity of input features is crucial.
For instance, action recognition frameworks that can only rely on optical-flow features, such as the work by Gao \etal~\cite{gao2018im2flow}, are more likely to fail than our approach, due to an excess of information in the input which is not related to the action itself.
In our work, we showed how the combination of classifiers relying on inputs with different context and meaning can significantly boost the accuracy, compensating for actions for which a method is less suitable and performs poorly. 
\section{Future Work}
The versatility and modularity of our approach allows it to be used for multiple purposes. 
As we employed our DA-PoTion on human joints for the task of full-body action recognition, it would be easy to adapt our framework on human hands to recognize fine-grained actions like playing piano, grasping objects, or pouring milk~\cite{yuan2018depth, simon2017hand, gattupalli2018towards, agarwal2017facial}.
Analogously, it might be interesting to further tune our approach for facial expression recognition. It would be as simple as replacing the 3D full-body pose regressor with a human face keypoint tracker~\cite{Sun_2013_CVPR, berretti20113d, zhang2016facial}.

Furthermore, given the DA-PoTion resulting from one section of a videoclip, one could train a Recurrent Neural Network (RNN) model to recursively predict future poses of the subject in the scene.

Despite our framework could easily be applied to multiple subjects simultaneously, current state-of-the-art 3D human pose predictors process only one subject at a time~\cite{zhou2017towards, pavlakos2017coarse, rhodin2018unsupervised, sun2018integral}.
Our work will therefore greatly benefit, in terms of general applicability, from advancements of 3D human pose regressors in that direction, just like the algorithm developed by Cao~\etal~\cite{cao2017realtime} did for the 2D setting.

{\small
\bibliographystyle{ieee}
\bibliography{egbib}
}

\end{document}